\documentclass[letterpaper, 10 pt, conference]{ieeeconf}  

\IEEEoverridecommandlockouts  

\usepackage{graphicx} 
\usepackage{amsmath}  
\usepackage{amssymb}  

\usepackage{cite}
\usepackage{booktabs}
\usepackage{multirow}
\usepackage{makecell}
\usepackage{pifont}
\usepackage{lscape}
\usepackage{url}
\usepackage{capt-of}
\usepackage[table]{xcolor}
\usepackage{subcaption}

\usepackage{url}
\usepackage[hidelinks]{hyperref}

\usepackage{caption}
\DeclareFontShape{OT1}{ptm}{m}{scit}{<->ssub*ptm/m/it}{}
\DeclareFontShape{OT1}{ptm}{b}{scit}{<->ssub*ptm/b/it}{}

\definecolor{ForestGreen}{HTML}{468432}

\newcommand{\todo}[1]{}

\title{\LARGE \bf
TrafficSignBench: Rule-Centric Closed-Loop Evaluation of Traffic-Sign Compliance in Autonomous Driving
}

\author{
Victoria Smirnova$^{1,*}$,
Viktoriia Zinkovich$^{2,*}$,
Gregorii Bukhtuev$^{3}$,
Artem Belyaev$^{1,4}$, \\
Andrey Kuznetsov$^{2,5}$,
Denis Shepelev$^{2,6,\dagger}$,
Vlad Shakhuro$^{1,2,6,\dagger}$
\thanks{
$^{1}$Lomonosov Moscow State University;
$^{2}$FusionBrain Lab;
$^{3}$Moscow Institute of Physics and Technology;
$^{4}$HSE University;
$^{5}$Innopolis University;
$^{6}$NUST MISIS.
$^{*}$Equal contribution.
$^{\dagger}$Project leads.
Corresponding author:
\href{mailto:smirnovavs@my.msu.ru}{\texttt{smirnovavs@my.msu.ru}}.
}
}

\IEEEaftertitletext{%
\begin{center}
\includegraphics[width=\textwidth]{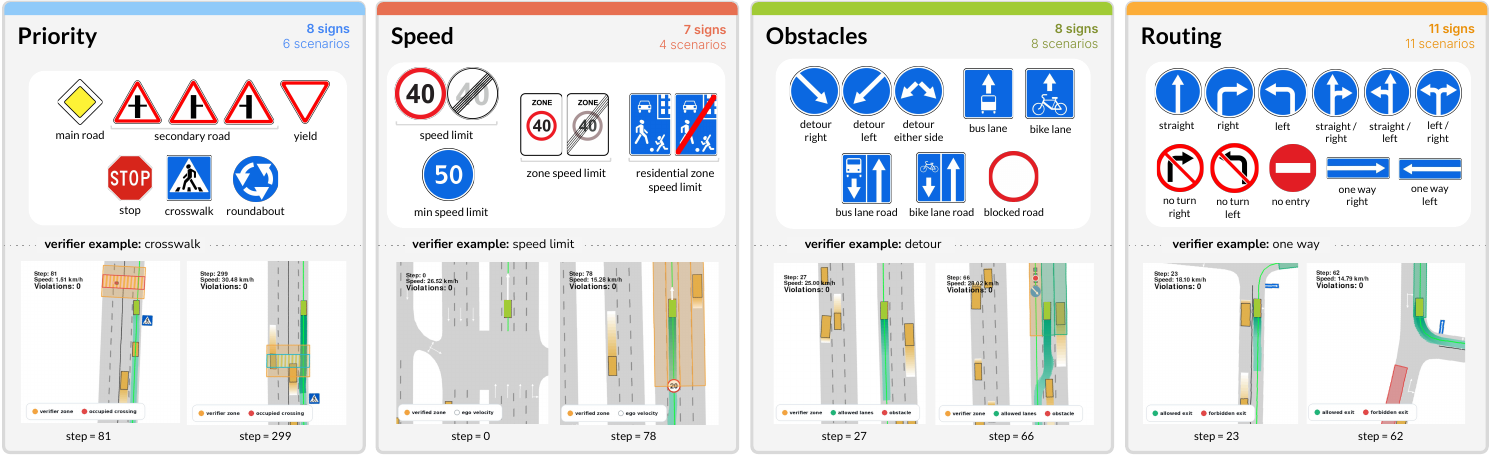}%
\end{center}%
\vspace{0.2em}%
\noindent
\refstepcounter{figure}%
Fig.~\thefigure.~~
\textbf{TrafficSignBench taxonomy of 34 formalized traffic signs across 4 semantic groups.}
For each semantic group, we present an example rule verifier implemented in our extended MetaDrive~\cite{li2022metadrive} simulator: (1)~\textit{Crosswalk} requires yielding to pedestrians; (2)~\textit{Speed Limit} enforces longitudinal deceleration to the posted limit; (3)~\textit{Detour} demands a lane change to bypass a stationary obstacle; and (4)~\textit{One-Way Road} requires strict compliance with mandatory direction.
\label{fig:taxonomy}%
\vspace{\baselineskip}%
}

\begin{document}

\maketitle
\thispagestyle{empty}
\pagestyle{empty}

\begin{abstract}

Autonomous driving planners are typically evaluated using aggregate metrics such as driving score, destination rate, and collision rate, which do not explicitly measure compliance with traffic rules. As a result, planners can achieve high benchmark scores while still exhibiting unsafe or illegal behaviors, limiting their applicability to real-world deployment. To address this gap, we introduce TrafficSignBench, a large-scale, traffic sign-centric benchmark for systematic and interpretable evaluation of traffic-rule compliance in autonomous driving. Our framework combines real-map-based simulation for realistic road layouts with rule-targeted procedural scenario generation for scalable and balanced coverage of underrepresented rules. We implement traffic rules corresponding to 34 traffic signs, each equipped with an automatic rule checker for detecting violations during closed-loop execution. This design yields 29,000 diverse road scenes and 29 distinct testing scenario types, enabling controlled evaluation of rule-specific planner behavior. We construct 5,800 testing scenes and demonstrate that current autonomous driving planners can exhibit poor traffic-rule compliance despite strong performance on standard evaluation metrics. To address this limitation, we transform existing planners into rule-compliant trajectory experts via explicit traffic-sign constraints, enabling scalable generation of high-quality oracle trajectories for fine-tuning.
Code and data are publicly available at \url{https://github.com/emb-ai/traffic-sign-bench} and \url{https://huggingface.co/datasets/emb-ai/traffic-sign-bench}.

\end{abstract}



\section{Introduction}


Autonomous driving (AD) systems are expected to operate safely and reliably in complex real-world traffic environments. 
Recent advances in learning-based driving planners have significantly improved trajectory generation in realistic traffic scenarios \cite{CARL, PLANT2, li2022BEVFormer, hu2023uniad}.
However, beyond safety and comfort, autonomous vehicles must also comply with traffic regulations that encode essential legal constraints of road behavior.

Despite this requirement, traffic-rule compliance remains largely underexplored in the evaluation of autonomous driving systems. 
Open-loop benchmarks \cite{wilson2023argoverse2generationdatasets, Ettinger_2021_ICCV, nuScenes} cannot capture violations by design, while closed-loop simulators and benchmarks \cite{dosovitskiy2017carlaopenurbandriving, bench2drive, nuPlan_dataset, commonroad} typically provide only partial coverage of traffic regulations and
rely primarily on aggregate safety, progress, and driving-performance metrics. 
Consequently, even recent benchmarks provide only limited and unsystematic coverage of rule compliance, allowing planners that achieve strong benchmark scores to still violate fundamental traffic regulations.

A key obstacle to systematic evaluation of traffic-rule compliance is the lack of suitable simulation infrastructure. 
While existing simulators provide realistic road layouts and interactive traffic participants \cite{dosovitskiy2017carlaopenurbandriving, rong2020lgsvlsimulatorhighfidelity, li2022metadrive}, explicit representations of traffic regulations are often limited: detecting rule violations usually requires manually designed scenarios or external evaluation tools.
Learning-based planners are typically trained on expert demonstrations \cite{bojarski2020nvidia, zheng2023diffusionplanner}, where traffic-rule compliance is learned implicitly and rule-specific corner cases are underrepresented. 
As a result, supervision on regulatory constraints remains limited, leading to poor generalization to rule-critical scenarios.

To address this gap, we introduce \textbf{TrafficSignBench}, an open-source benchmark for \emph{systematic and interpretable} evaluation of traffic-rule compliance in autonomous driving.

Overall, our contributions are as follows:

\begin{itemize}
    \item We extend MetaDrive~\cite{li2022metadrive} with executable rules for 34 traffic signs, each paired with an automatic online checker. A
    rule-targeted generation module creates interactions in which compliance
    requires an explicit planning decision.


    \item We introduce, to the best of our knowledge, the first large-scale benchmark to jointly combine a broad traffic-sign taxonomy, automatic rule checkers, and rule-targeted closed-loop scenarios for systematic traffic-rule compliance evaluation. TrafficSignBench contains 29,000 scenarios across 29 functional scenario types, constructed from a pool of 26,020 real-map crops. Scenario parameters are calibrated to real-world driving data while preserving controlled rule-critical interactions.
    
    \item We evaluate 17 planners and show that conventional
    metrics obscure systematic rule failures. Under a joint criterion that
    requires both target-rule compliance and destination completion, standard
    planners attain only 2.9--9.0\%. Fine-tuning PlanT-2 on
    oracle-selected expert trajectories raises this rate from 5.9\% to
    72.3\%, providing a strong baseline for rule-aware planning.
\end{itemize}

\section{Related Work}

\subsection{Planning Benchmarks}

Autonomous-driving benchmarks can be divided into open-loop and closed-loop evaluation approaches.
Open-loop benchmarks \cite{wilson2023argoverse2generationdatasets, Ettinger_2021_ICCV} evaluate models via trajectory displacement error relative to expert behavior without executing the planner, making traffic rule violations unobservable by design.
Closed-loop benchmarks address this limitation by executing models in simulation \cite{dosovitskiy2017carlaopenurbandriving, bench2drive, nuPlan_dataset}. 
Yet log-derived simulators often lack a queryable traffic-control layer: most signs and road markings are neither annotated nor modeled as simulator objects. 
As a result, they primarily report generic safety, comfort, and progress metrics \cite{nuPlan_dataset, Waymax_bench}. 
The NavSim benchmark \cite{navsim_bench} links open- and closed-loop evaluation through non-reactive BEV unrolling, but its unified metric is too coarse for fine-grained assessment and does not capture even basic rule compliance. 
Bench2Drive \cite{bench2drive} advances closed-loop evaluation by introducing diverse interactive driving scenarios (e.g., cut-in, lane change) and explicit CARLA infraction detectors.
Its evaluation, however, is primarily organized around driving abilities and route-level performance rather than a systematic taxonomy of individual traffic-sign rules. CommonRoad~\cite{commonroad} represents traffic signs, and separate rule-monitoring work~\cite{maierhofer2022trafficrules} formalizes selected traffic rules, but neither provides a sign-centric benchmark with rule-targeted scenarios and per-rule planner evaluation.
TrafficSignBench addresses this gap by combining explicit traffic-sign and road-marking semantics, deterministic per-rule compliance checkers, and rule-targeted scenario generation for systematic rule-centric evaluation. 
\subsection{Simulators for Autonomous Driving}
Evaluation of autonomous driving models is broadly conducted in either sensor \cite{dosovitskiy2017carlaopenurbandriving, rong2020lgsvlsimulatorhighfidelity, ito2025dawsim, lesy2025asvsim} or data-driven simulation \cite{li2022metadrive, nuPlan_dataset, Waymax_bench}. 
These paradigms differ in whether they focus on reproducing realistic sensor observations or enabling scalable evaluation of decision-making planners \cite{navsim_bench}. 
In this work, we focus on the latter class, where methods typically operate on semantic bird’s-eye-view (BEV) representations \cite{li2022BEVFormer} that support large-scale scenario generation and planner evaluation independently of perception errors \cite{hu2023uniad, liao2025DiffusionDrive}. 
However, despite modeling road layouts and agent interactions, most existing simulators lack explicit representations of traffic rules, with evaluation often relying on manually designed scenarios. 
To address this limitation, we extend the MetaDrive simulator \cite{li2022metadrive} with explicit traffic-rule modeling, incorporating traffic signs from a real-world sign database and map data, together with automatic rule checkers for violation detection.


\subsection{Rule-Based and Learning-Based Planning}

Autonomous driving planners are commonly based on either rule-based or learning-based paradigms. 
Rule-based methods define driving behavior using manually designed rules and heuristics, offering strong interpretability and natural enforcement of safety constraints~\cite{dauner2023parting, vitelli2021safetynet, helbing1998generalized, hagedorn2025when}. 
However, their reliance on handcrafted logic limits scalability and generalization to complex or out-of-distribution scenarios.
Learning-based methods instead learn policies from data via imitation~\cite{zheng2023diffusionplanner, PLANT2, cheng2024pluto} or reinforcement learning~\cite{CARL, li2024think2driveefficientreinforcementlearning}. 
While they achieve strong performance in data-driven settings, they do not explicitly encode traffic regulations in the training objective. 
As a result, rule compliance is typically enforced only implicitly or through post-hoc trajectory filtering. 
This increases inference complexity and can introduce failure modes under distribution shift, where rule violations are harder to detect and correct.

Overall, rule-based systems are interpretable but not scalable, whereas learning-based methods are flexible but lack explicit rule adherence and safety guarantees. 
These limitations motivate the need for a large-scale benchmark for the systematic evaluation of traffic rule compliance in modern planners.

\section{TrafficSignBench}
\label{sec:overview}

In this section, we introduce TrafficSignBench, a benchmark designed to evaluate how autonomous driving planners comply with the rules imposed by traffic signs.
First, we formalize a core set of traffic rules and validate their cross-country generalizability (Section~\ref{sec:rule_formalization}). 
Second, we organize these rules into a functional testing taxonomy based on the semantic capabilities required from the ego vehicle (Section~\ref{sec:taxonomy}). 
Third, we construct a diverse testing environment by procedurally placing these signs onto a large pool of real-world map crops (Section~\ref{sec:real-map}). 
Finally, we generate rule-targeted, closed-loop scenarios calibrated with real-world traffic statistics to ensure that compliance requires explicit, non-trivial planning decisions (Section~\ref{sec:rule-scenario}).

\subsection{Rule Formalization and Generalizability}
\label{sec:rule_formalization}

Despite regional variation, traffic signs and their associated rules largely follow shared international conventions. 
We adopt a representative European traffic code—Russian traffic rules—as our reference, as they conform to the \textit{Vienna Convention on Road Signs and Signals} (1968). 
Traffic signs vary in function, and not all are suitable for systematic evaluation. 
We therefore consider signs that impose explicit, mandatory rules on road users, while excluding warning or informational signs whose semantics primarily provide contextual or conditional information (e.g., time- or vehicle-specific restrictions). 
To further ensure practical relevance, we cross-reference the resulting set with the Moscow traffic sign database~\cite{moscow_signs} and discard signs not observed in practice. 
This procedure results in a core set of \textbf{34 traffic signs} (Figure~\ref{fig:taxonomy}), all of which we implement in the simulator, each with explicitly defined and verifiable rule logic.

To validate that this core set generalizes beyond our reference system, we analyzed its cross-country semantic alignment. 
For each country, we analyze whether each of the 34 implemented signs has a \textit{semantic equivalent}—a sign (or combination of signs and/or road markings) that encodes the same driving rule, regardless of visual differences. Table~\ref{tab:country_comparison} reports the match rate per sign category as well as the overall mean.
Our results show that adapting the benchmark to other countries requires small modifications, with an average semantic overlap of \textbf{92\%} across Vienna Convention signatories and substantial alignment even in non-Vienna systems.
Consequently, the TrafficSignBench modular design enables adaptation to other jurisdictions by simply substituting the visual appearance of signs without modifying the underlying rule-verification logic.

\begin{table}[ht]
\centering
\scriptsize
\setlength{\tabcolsep}{5pt}
\renewcommand{\arraystretch}{1.15}

\begin{tabular}{lcccccc}
\toprule
& \multicolumn{4}{c}{\textbf{Vienna Signatories}} 
& \multicolumn{2}{c}{\textbf{Non-Vienna}} \\
\cmidrule(lr){2-5} \cmidrule(lr){6-7}

&
\makecell[c]{\includegraphics[width=0.25cm]{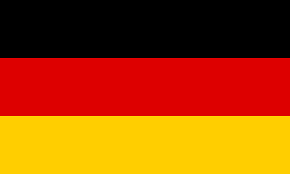} Germany\\[-1pt]
\cite{germany_road_signs}}
&
\makecell[c]{\includegraphics[width=0.25cm]{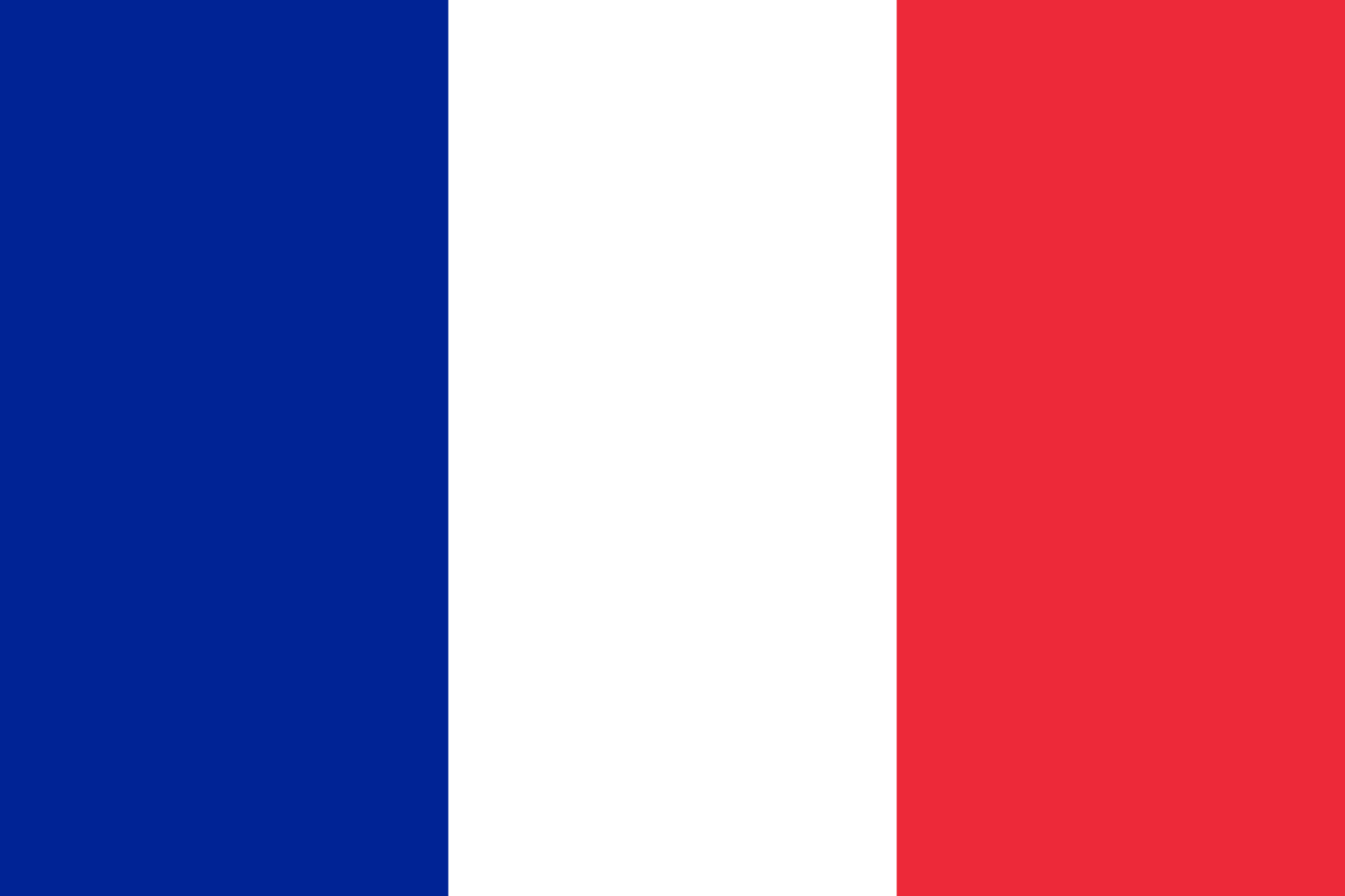} France\\[-1pt]
\cite{france_road_signs}}
&
\makecell[c]{\includegraphics[width=0.25cm]{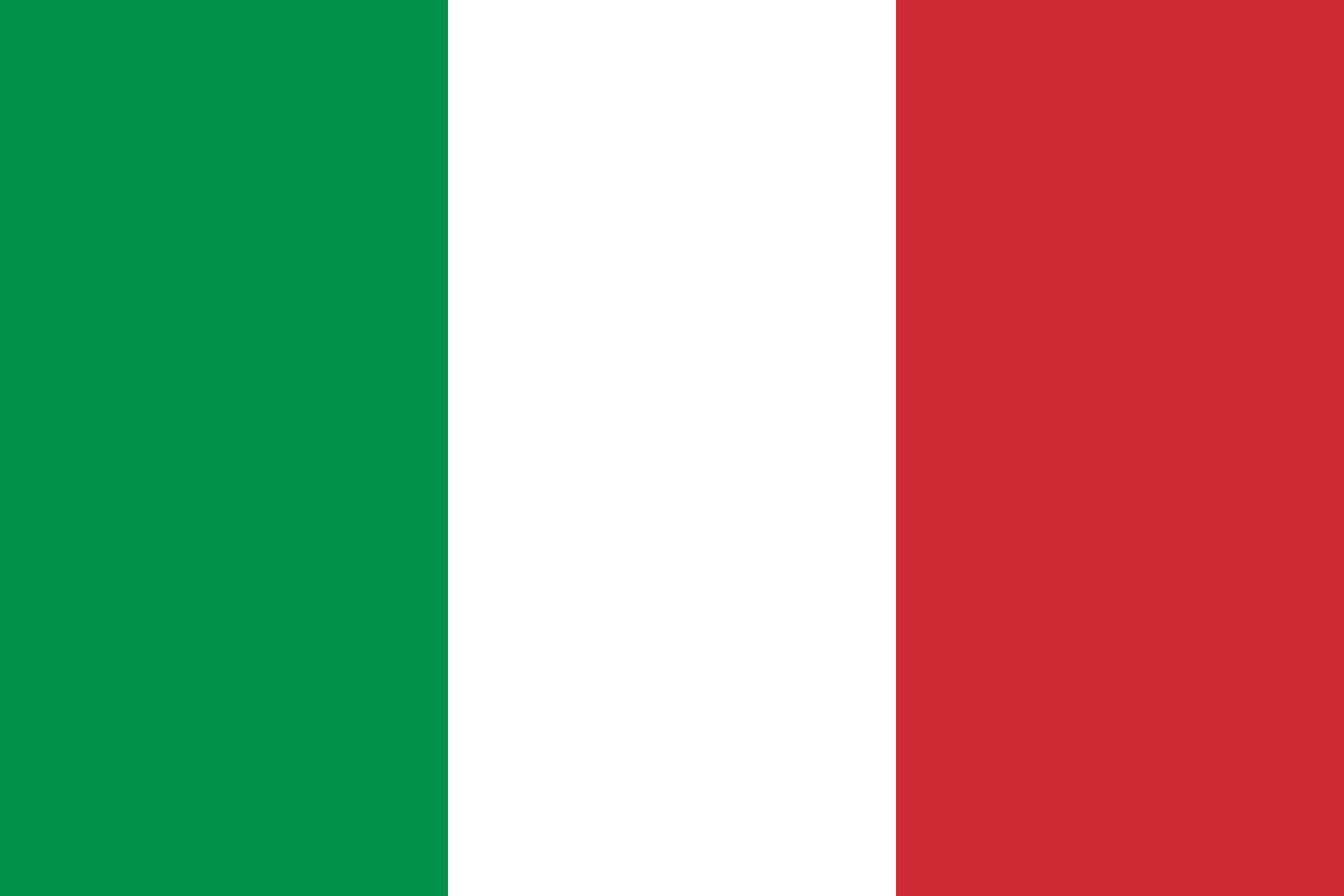} Italy\\[-1pt]
\cite{italy_road_signs}}
&
\makecell[c]{\includegraphics[width=0.25cm]{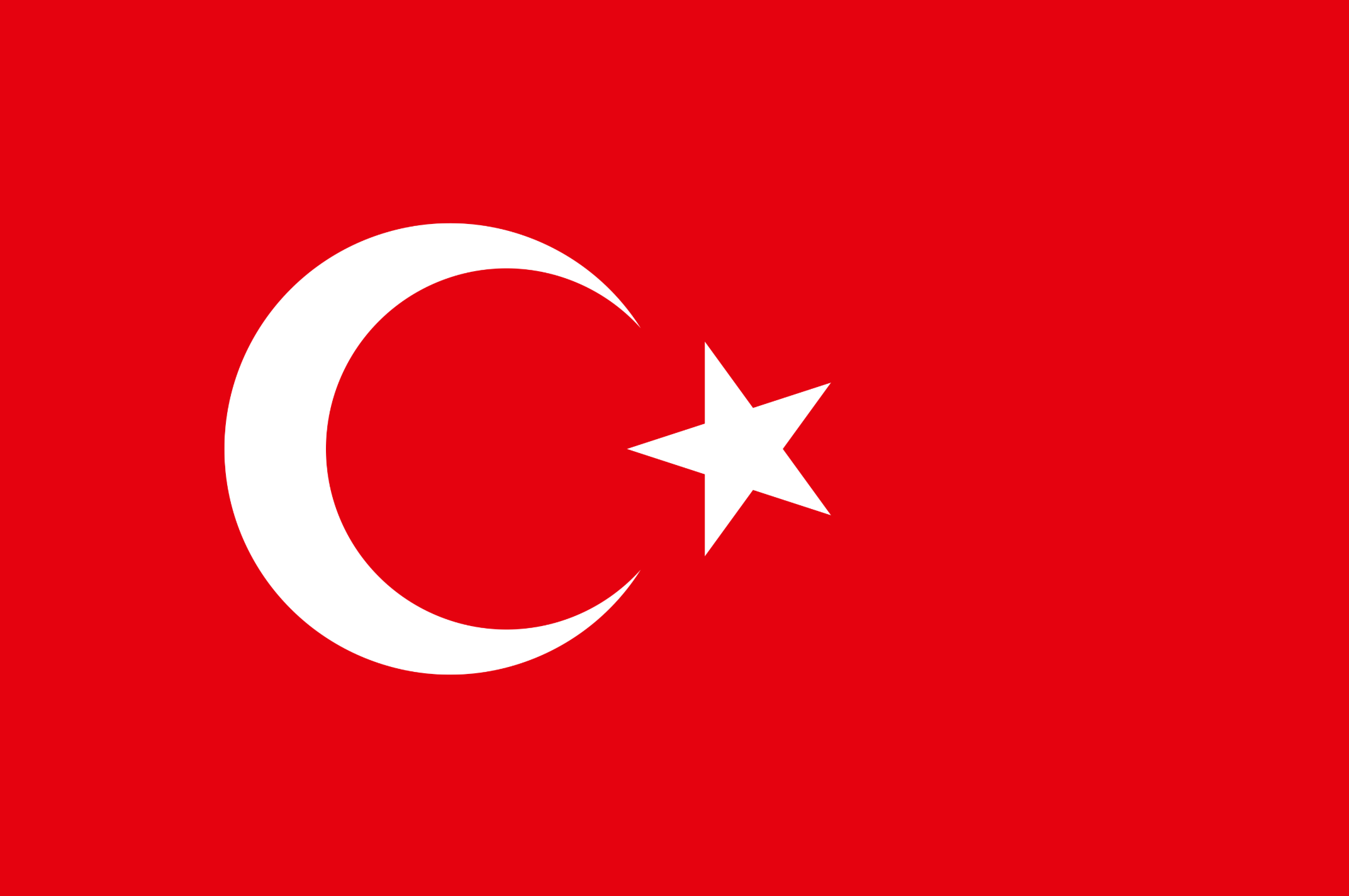} Turkey\\[-1pt]
\cite{turkey_road_signs}}
&
\makecell[c]{\includegraphics[width=0.30cm]{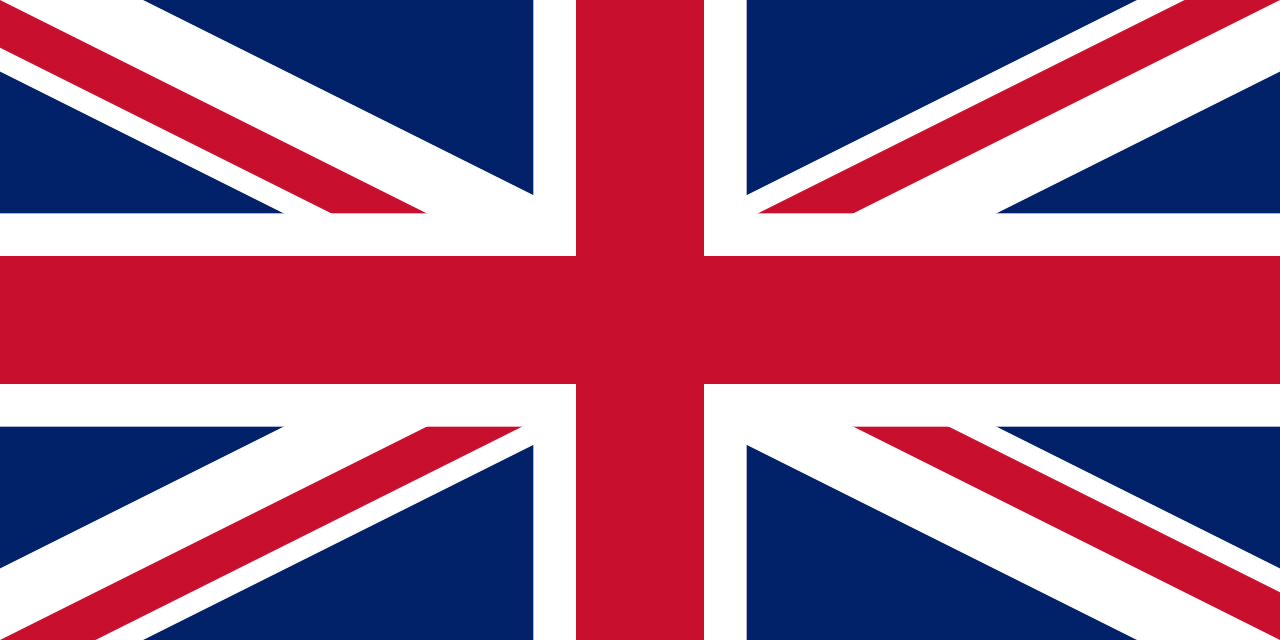} UK\\[-1pt]
\cite{uk_highway_code}}
&
\makecell[c]{\includegraphics[width=0.25cm]{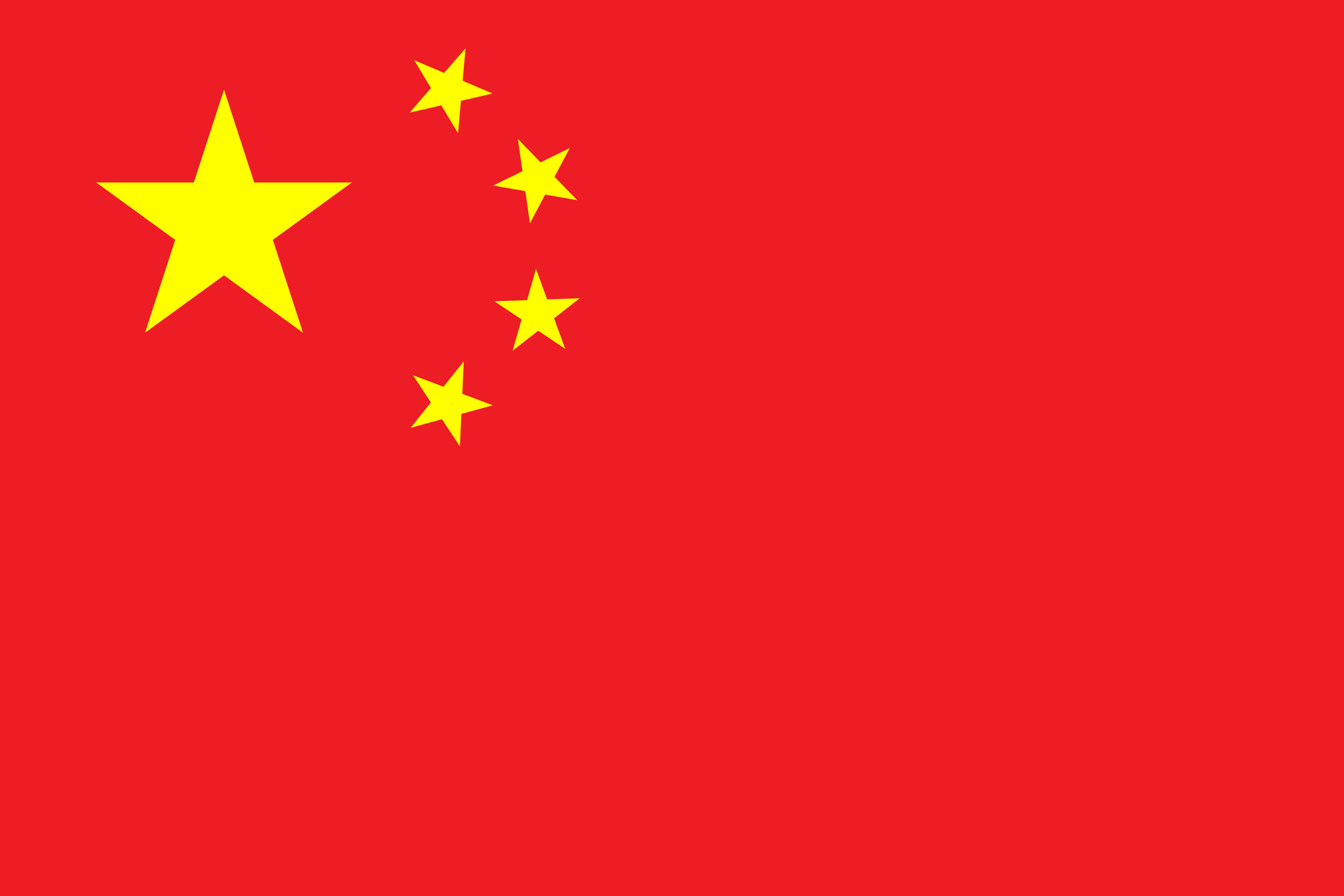} China\\[-1pt]
\cite{china_road_signs}}
\\
\midrule

\textbf{Mean} &
91\% &
84\% &
94\% &
100\% &
97\% &
91\% \\

\bottomrule
\end{tabular}

\caption{
\textbf{Semantic equivalence of TrafficSignBench signs across countries.}
Values report the percentage of the 34 implemented signs that have a semantic equivalent with Russian traffic signs~\cite{russia_pdd_2025}.
}
\label{tab:country_comparison}
\end{table}

\subsection{Testing Taxonomy}
\label{sec:taxonomy}

TrafficSignBench formalizes rules for 34 traffic signs, but these signs do not all correspond to distinct closed-loop testing scenarios. 
Namely, three signs terminate an active restriction (e.g., end of speed limit) and are paired with the rules they conclude, while three secondary-road signs share identical legal effects across different map geometries. 
Accounting for these functional overlaps, we consolidate the signs into \textbf{29} distinct closed-loop scenario types. 

As detailed in Figure~\ref{fig:taxonomy}, we organize these scenarios into \textbf{4 semantic groups} based on the high-level capability demanded of the ego vehicle:
(i)~\textbf{Priority} (6 scenarios) tests right-of-way reasoning and interactions with conflicting agents at unsignalized intersections, roundabouts, and crosswalks;
(ii)~\textbf{Speed} (4 scenarios) evaluates longitudinal velocity control under maximum, minimum, and area-wide limits;
(iii)~\textbf{Obstacles} (8 scenarios) requires safe navigation around inaccessible space, including stationary detours, blocked roads, and reserved bus or bicycle lanes; and (iv)~\textbf{Routing} (11 scenarios) requires the ego vehicle to avoid prohibited road segments or maneuvers and adapt its route according to current traffic restrictions, covering no-entry zones, prohibited turns, mandatory
directions, and one-way streets.

\subsection{Real-Map Scene Construction}
\label{sec:real-map}

Closed-loop rule tests on synthetic layouts can reward fitting the generator rather than reading the sign.
TrafficSignBench instead harvests \emph{sign-free} lane-level fragments from real-world Moscow maps via OpenStreetMap~\cite{openstreetmap, Enhancing_SUMO}.
The evaluated signs are then procedurally placed onto these real-world crops based on their semantic requirements (e.g., priority signs are exclusively placed at junctions). 
This approach yields a diverse pool of \textbf{26{,}020} real-world crops in three structural families: 
(i)~junctions (6{,}457 T/X intersections and roundabouts) for priority scenarios; 
(ii)~dual-path maps (6{,}507) for routing scenarios, which pair a shorter prohibited route with a longer compliant route to the same destination, explicitly testing the planner's ability to prioritize rule compliance over route efficiency; and 
(iii)~corridors (13{,}056) for speed, obstacle, and crosswalk tests.
To guarantee structural variety, we sample roads across different lane counts and curvature profiles. 
This approach ensures broad geographic diversity while preserving realistic map topologies and valid sign placements.

Maps are split $80$/$20$ \emph{before} sign assignment by unique ID from OpenStreetMap to exclude train/test data leakage.
Each sign then receives \textbf{100} maps (\textbf{80} train / \textbf{20} test) from its compatible family, with balanced T/X or curvature/lane quotas.

\subsection{Rule-Targeted Scenario Generation}
\label{sec:rule-scenario}

Random placement of the ego vehicle and surrounding agents rarely probes the intended rule: conflicting traffic may never appear, or the posted limit may already match the ego's default speed. 
To guarantee meaningful evaluation, we explicitly design \emph{rule-targeted} interactions. 
For example, for yielding scenarios, we introduce \emph{task agents}—a gated convoy on the main road that is released exactly as the ego approaches the conflict zone. 
On dual-path maps, the ego is assigned a destination that makes the prohibited branch the shortest path, explicitly tempting the planner to violate the rule. 
For crosswalks, pedestrians follow controlled temporal presets, including late-arriving individuals and chained groups.

\begin{figure}[th]
    \centering
    \includegraphics[width=\linewidth]{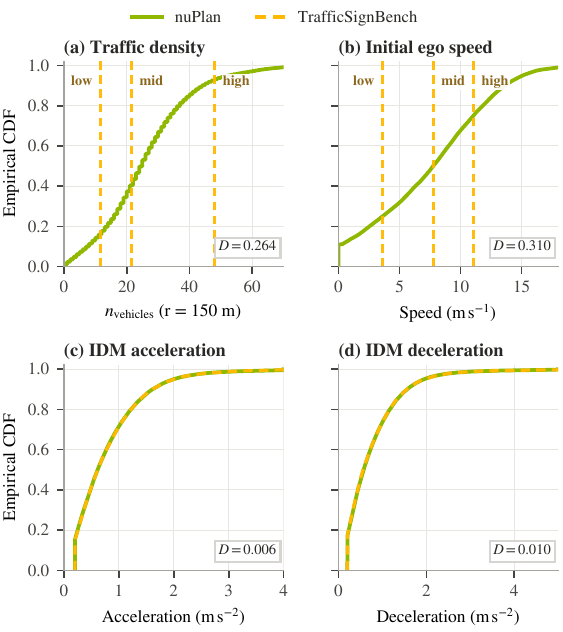}
    \caption{
        \textbf{Calibration of background traffic dynamics.} 
        Empirical CDFs compare real-world data from the nuPlan dataset~\cite{nuPlan_dataset} with values recorded in
        TrafficSignBench; dashed guides mark the three
        controlled density and speed probes, and \(D\) is the two-sample KS
        statistic.
    }
    \label{fig:nuplan_stats}
\end{figure}

To ensure statistical significance and robust evaluation, we expand each of the 2,900 maps into \textbf{10 distinct scenarios} by introducing variability across 5 key dimensions: ego spawn lane, route length, initial speed, traffic density, and background agent dynamics. 
Background agents are controlled using the Intelligent Driver Model (IDM) \cite{IDM}, a standard widely adopted rule-based policy for reactive background agents \cite{navsim_bench, li2022metadrive, Enhancing_SUMO}, which produces realistic behavior.
To maintain realism, all stochastic parameters (initial speed, density, and IDM parameters) are sampled from distributions empirically estimated from the nuPlan dataset~\cite{nuPlan_dataset} (Fig.~\ref{fig:nuplan_stats}).
Density and initial speed are deliberately drawn from a small set of nuPlan
quantiles, so that each functional
group can be probed
under sparse/dense and slow/fast conditions (Fig.~\ref{fig:nuplan_stats}a--b).

In total, with 29 distinct testing scenario types instantiated across 100 real-map crops (80 train / 20 test) and expanded into 10 variants each, TrafficSignBench yields \textbf{29{,}000} closed-loop scenarios (\textbf{23{,}200} train / \textbf{5{,}800} test). 
This scale provides a rigorous, diverse, and statistically robust foundation for systematic planner evaluation.





\section{Evaluation Protocol}
\label{sec:evaluation}

\subsection{Rule-Aware Evaluation Metrics}
\label{sec:metrics}

Standard metrics, used in established evaluation protocols~\cite{bench2drive, dosovitskiy2017carlaopenurbandriving, nuPlan_dataset}, do not explicitly measure compliance with traffic regulations.  
A planner can achieve near-perfect scores on conventional benchmarks while systematically violating traffic rules.
To address this gap, we complement standard metrics (Driving Score, Destination Rate, Efficiency, and Collision Rate) with explicit rule-adherence measurements derived from our automatic rule-based verifiers.

Our primary metric is \textbf{Sign Compliance $\times$ Destination (SCD)}, where Sign Compliance (SC) shows whether target-sign verifier records a violation. 
Evaluating sign compliance in isolation can be ambiguous: if an agent enters a sign's zone but crashes or gets stuck before completing the maneuver (e.g., failing to finish a turn), it is impossible to definitively determine whether the rule would have been fully respected. 
To provide both high-level summaries and fine-grained interpretability, we report SCD at three levels of aggregation:
\begin{itemize}
    \item \textbf{SCD}: The baseline metric computed individually for each of the 29 scenario types, enabling granular analysis of planner behavior on specific traffic signs.
    \item \textbf{Group SCD}: The metric aggregated within each of the four semantic groups (\textit{Priority}, \textit{Speed}, \textit{Obstacles}, and \textit{Routing}), highlighting specific operational failure modes.
    \item \textbf{Overall SCD}: The macro-average across all 29 scenario types, providing a single summary statistic of a planner's rule-following capability.
\end{itemize}

\subsection{Baselines}
\label{sec:baselines}

We use TrafficSignBench to systematically evaluate rule compliance across a diverse set of self-driving planners. 
For clarity, we group the \textbf{17 evaluated planners} into 4 categories based on their access to explicit traffic rules:

\textbf{(1) Base Planners.} These pre-trained methods operate without explicit access to traffic rules, relying solely on implicit learning or predefined behavioral heuristics:
\begin{itemize}
    \item \textbf{IDM}~\cite{IDM}: A deterministic rule-based car-following model. We evaluate the default IDM alongside four stochastic variants (IDM-$s_1$ to IDM-$s_4$) with parameters sampled from data-driven priors.
    \item \textbf{PPO}~\cite{li2022metadrive}: A standard reinforcement learning planner trained with Proximal Policy Optimization.
    \item \textbf{CaRL}~\cite{CARL}: A reinforcement learning planner trained with sparse route-completion rewards.
    \item \textbf{PlanT-2}~\cite{PLANT2}: A state-of-the-art transformer-based imitation learning planner.
\end{itemize}

\textbf{(2) Rule-Compliant Experts.} For each base planner, we construct its corresponding rule-compliant expert (denoted with an $e$ superscript, e.g., $\text{IDM}^{e}$, $\text{PlanT-2}^{e}$). These experts enforce explicit, rule-based action constraints (e.g., hard braking before a stop line) within regions governed by traffic signs, without modifying the underlying model architecture. \todo{Add more details about ifs.}

\textbf{(3) Oracle Expert.} We define an upper-bound oracle by aggregating all rule-compliant experts and selecting the best-performing one for each scene. 
This oracle is further used for collection of ground-truth trajectories (see Section~\ref{sec:trajectory}).Since it performs post-hoc selection over multiple expert rollouts, it is not included as an evaluation baseline.

\textbf{(4) Rule-Supervised Baseline.} As an additional learning-based baseline, we evaluate \textbf{PlanT-2-FT}, a supervised variant of PlanT-2 fine-tuned exclusively on trajectories generated by the Oracle Expert. Details on its architecture and training are provided in Section~\ref{sec:finetuning}.


\begin{figure}[ht]
    \centering
    \includegraphics[width=\linewidth]{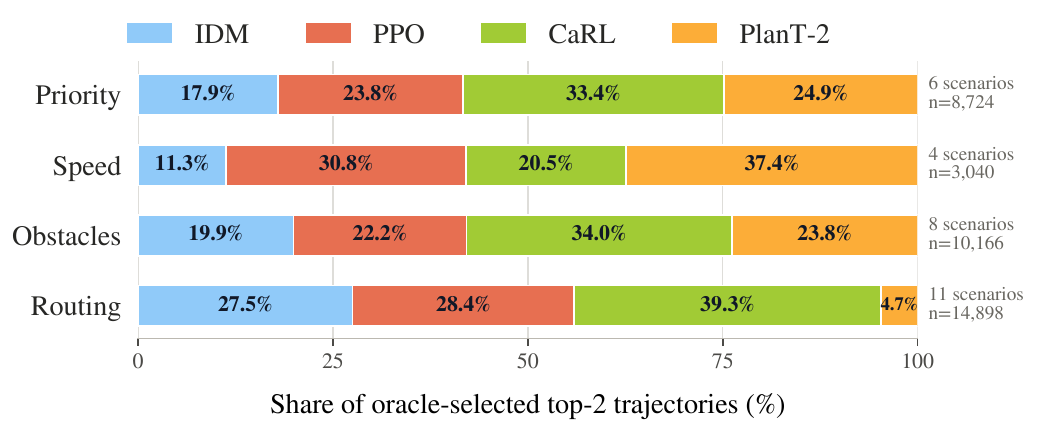}
    \caption{
    \textbf{Composition of the Oracle Expert dataset.} 
    Proportion of high-quality trajectories contributed by each of the 8 rule-compliant experts across the 4 semantic sign groups. 
    The diverse distribution shows that no single planner dominates; rather, the experts exhibit complementary strengths, providing robust and varied demonstrations.
    }
    \label{fig:trajs}
\end{figure}

\subsection{Oracle Trajectory Collection}
\label{sec:trajectory}

We collect expert trajectories for 23,200 training scenarios as follows.
For each scenario, we run all \textbf{8 expert planners} (Section~\ref{sec:baselines}) on the same scenario initialization, which allows their trajectories to be compared under matched closed-loop conditions.
For each trajectory, we compute the metrics described in Section~\ref{sec:metrics}.
A trajectory is considered successful only if the ego vehicle completes the route without violating the target traffic rule (i.e., $\text{SCD} = 100\%$).

If an expert fails under the shared initialization, we iteratively resample the scenario configuration for that specific expert until a successful trajectory is achieved, ensuring valid demonstrations even in challenging edge cases.
To construct the final oracle trajectories collection, we apply a rigorous filtering and ranking procedure. 
First, we isolate all successful trajectories. 
Among the IDM-based experts, we retain only the fastest rollout, as these controllers are consistently smooth and differ primarily in efficiency. 
We then compare this IDM trajectory against all successful non-IDM experts using a weighted F-score that balances normalized speed $\tilde{v}(\tau)$ and comfort $\tilde{s}(\tau)$:
\begin{equation*}
Q(\tau) =
\frac{(1+\beta^2)\,\tilde{v}(\tau)\,\tilde{s}(\tau)}
{\beta^2\,\tilde{v}(\tau) + \tilde{s}(\tau)},
\qquad \beta = 0.25,
\end{equation*}
where $\beta < 1$ prioritizes driving efficiency over comfort. 
For each training scenario, we retain the top-2 trajectories according to $Q(\tau)$, resulting in a final dataset of \textbf{36,828 high-quality training trajectories} (on average 1.6 successful per scenario). 
As shown in Figure~\ref{fig:trajs}, different expert planners contribute substantially across various sign categories, demonstrating complementary strengths rather than a single dominant policy.

\subsection{Rule-Supervised Fine-Tuning}
\label{sec:finetuning}

\textbf{Fine-tuning PlanT-2.}\enspace
PlanT-2-FT fine-tunes a \(37.2\,\mathrm{M}\)-parameter PlanT-style HFLM planner on oracle trajectories from the training split. The backbone is unchanged; traffic signs are encoded as object tokens with learned class projections. A persistent sign-state token stores the active sign class and posted value, preserving zone-level restrictions. A learnable speed token feeds a discrete ego-speed head trained jointly with the path and waypoint heads. These extensions add 147{,}976 parameters (0.4\%).

\textbf{Training.}\enspace
We jointly optimize \(L_1\) path and waypoint losses and cross-entropy over soft two-hot speed targets. Training uses 29{,}462 trajectories from the 80\% training split, while the remaining 7{,}366 trajectories are used for validation, across all 29 scenario types. We fine-tune for 26 epochs using AdamW, cosine decay, a \(10^{-4}\) base learning rate, gradient clipping at 1.0, and batch size 128 on 1 NVIDIA A100. The backbone uses layer-wise learning-rate decay of 0.8, while new parameters use a \(5\times\) multiplier; new sign projections are initialized from the closest pretrained class. Checkpoints are selected by closed-loop validation performance.

\textbf{Training supervision.}\enspace
Speed-regime targets use the minimum expert speed over the next \(4\,\mathrm{s}\); for minimum-speed sign 4.6, the target is \(3\,\mathrm{km/h}\) above the posted value. Maneuver signs use the expert \(4\,\mathrm{s}\) trajectory as path target rather than the input route. For travel-direction restrictions, the route follows the prohibited branch, forcing sign-conditioned maneuver selection. Right-of-way signs use frame-local targets. We up-weight sign-relevant decision windows and balance sign families such that none exceeds 25\% of a batch. Finally, each frame is additionally rendered from an ego pose perturbed by \(1.0\,\mathrm{m}\) laterally and \(5^\circ\) in yaw, while retaining the original trajectory targets to supervise recovery onto the demonstrated track.














\section{Experimental Results}
\label{sec:results}

\begin{table*}[t]
\centering
\small
\begin{tabular*}{\textwidth}{@{\extracolsep{\fill}}lcccc@{\hspace{8pt}}ccccc@{}}
\toprule
& \multicolumn{4}{c}{\textbf{Conventional driving}}
& \multicolumn{4}{c}{\textbf{Group SCD (\%)}} \\
\cmidrule(lr){2-5}\cmidrule(lr){6-9}
\textbf{Planner}
& \makecell{\textbf{DS} $\uparrow$}
& \makecell{\textbf{Dest.} (\%) $\uparrow$}
& \makecell{\textbf{Eff.} $\uparrow$}
& \makecell{\textbf{Coll.} (\%) $\downarrow$}
& \makecell{\textbf{Priority} $\uparrow$}
& \makecell{\textbf{Speed} $\uparrow$}
& \makecell{\textbf{Obstacle} $\uparrow$}
& \makecell{\textbf{Routing} $\uparrow$}
& \makecell{\textbf{Overall}\\SCD $\uparrow$} \\
\midrule
\multicolumn{10}{l}{\textit{Standard baselines}} \\[-1pt]
IDM & 35.7 & 66.4 & 115.1 & 21.0 & 14.6 & 26.6 & 1.7 & 0.0 & 7.2 \\
IDM-$s_1$ & 27.6 & 51.8 & 84.3 & 19.2 & 17.4 & 34.6 & 2.2 & 0.0 & 9.0 \\
IDM-$s_2$ & 28.0 & 52.3 & 84.0 & 19.3 & 17.9 & 33.6 & 2.3 & 0.0 & 9.0 \\
IDM-$s_3$ & 27.1 & 51.0 & 84.7 & 20.2 & 18.6 & 32.0 & 1.9 & 0.0 & 8.8 \\
IDM-$s_4$ & 27.4 & 51.6 & 84.5 & 20.0 & 18.5 & 33.0 & 1.8 & 0.0 & 8.9 \\
PPO & 35.6 & 67.9 & 151.1 & 27.5 & 3.0 & 17.0 & 0.9 & 0.0 & 3.2 \\
CaRL & 39.6 & 73.7 & 149.4 & 21.0 & 2.0 & 10.2 & 3.7 & 0.1 & 2.9 \\
PlanT-2 & 28.6 & 56.1 & \textbf{208.4} & 43.3 & 5.8 & 29.8 & 0.7 & 1.1 & 5.9 \\
\noalign{\vskip 2.5pt{\color{black!35}\hrule height 0.35pt}\vskip 2.5pt}
\textbf{PlanT-2-FT} & \textbf{74.2} & \textbf{74.8} & 99.9 & \textbf{12.0} & \textbf{57.2} & \textbf{96.2} & \textbf{90.8} & \textbf{58.4} & \textbf{72.3} \\
\midrule
\multicolumn{10}{l}{\textit{Privileged rule-compliant baselines}} \\[-1pt]
IDM$^{e}$ & 78.2 & 79.2 & 110.0 & 15.9 & 81.2 & 94.8 & 88.3 & 59.7 & 76.9 \\
IDM$^{e}$-$s_1$ & 65.4 & 66.0 & 83.1 & 10.9 & 68.2 & 86.2 & 79.6 & 44.3 & 64.8 \\
IDM$^{e}$-$s_2$ & 65.5 & 66.2 & 83.0 & 11.2 & 68.9 & 86.2 & 76.9 & 45.3 & 64.6 \\
IDM$^{e}$-$s_3$ & 65.3 & 65.9 & 83.7 & 11.3 & 69.2 & 86.2 & 76.9 & 44.6 & 64.3 \\
IDM$^{e}$-$s_4$ & 65.7 & 66.2 & 83.4 & 10.9 & 68.5 & 87.8 & 77.8 & 44.9 & 64.8 \\
PPO$^{e}$ & 78.7 & 79.6 & 122.3 & 16.9 & 77.3 & 94.4 & 86.9 & 64.4 & 77.4 \\
CaRL$^{e}$ & 80.9 & 81.3 & 122.0 & 13.0 & 80.6 & 95.2 & 87.9 & 68.5 & 80.0 \\
PlanT-2$^{e}$ & 69.2 & 70.7 & 156.1 & 28.5 & 82.6 & 98.0 & 79.6 & 38.8 & 67.3 \\
\bottomrule
\end{tabular*}
\caption{\textbf{Closed-loop evaluation on all 29 scenario types.}
DS: Driving Score; Dest.: Destination Rate; Eff.: Efficiency; Coll.: Collision Rate.
Conventional metrics are episode-weighted; SCD is macro-averaged over scenario types, giving every rule equal weight.
Bold denotes the best result among standard baselines and our method.
Superscript $e$ denotes privileged rule-compliant policies with explicit access to the target rule; they are reported as privileged reference policies.
Confidence intervals are shown in Fig.~\ref{fig:metrics_CI} and omitted here for readability.}
\label{tab:metrics}
\end{table*}

\begin{figure*}[t]
    \centering
    \includegraphics[width=\linewidth]{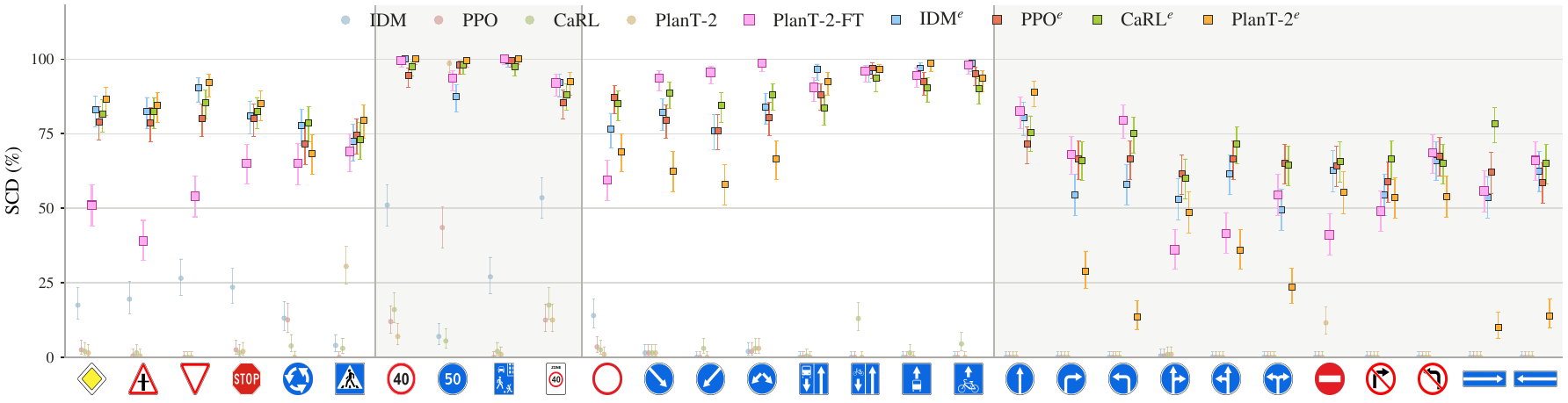}
    \caption{
    \textbf{Overall rule-conditioned performance across 29 scenario types.}
    Points show individual SCD; whiskers show the sample confidence interval across scenario-level
    rates. 
    Superscript $e$ denotes policies
    with privileged access to the target rule.
    }
    \label{fig:metrics_CI}
\end{figure*}

\subsection{Planner Comparison}
\label{sec:planner_comparison}

TrafficSignBench exposes a near-total failure of standard planners on explicit
rules: Table~\ref{tab:metrics} reports only 2.9--9.0\% overall SCD.
Rule-targeted supervision changes the operating regime. PlanT-2-FT reaches
72.3\%, a 66.4-point gain over its PlanT-2 initialization and a 63.3-point
gain over the strongest standard baseline. The improvement is not confined to
one rule family: Obstacle rises from 3.7\% to 90.8\%, while Routing rises
from at most 1.1\% to 58.4\%.

The benchmark is therefore difficult, but not intrinsically unsolvable.
PlanT-2-FT comes within 7.7 points of the strongest privileged expert
(80.0\%) and surpasses every expert on Obstacle scenarios. The remaining gap
is sharply localized to \textbf{Priority} and \textbf{Routing}, identifying interactive
right-of-way and sign-conditioned route selection as the central unsolved
behaviors. Figure~\ref{fig:metrics_CI} summarizes this separation.

\begin{table}[ht]
\caption{
\textbf{Sign-channel ablation.}
PlanT-2-FT evaluated on paired scenes with sign identity removed from the input; plates remain as
anonymous static objects. Parentheses indicate the number of sign families; $n$ is the number of
paired episodes.
}
\centering
\setlength{\tabcolsep}{4pt}
\renewcommand{\arraystretch}{1.25}
\begin{tabular}{lccccc}
\toprule
& \multicolumn{1}{c}{}
& \multicolumn{2}{c}{\textbf{SCD} (\%) $\uparrow$}
& \multicolumn{2}{c}{\textbf{Sign Compliance} (\%) $\uparrow$} \\
\cmidrule(lr){3-4} \cmidrule(lr){5-6}
\textbf{Sign group}
& \makecell{$n$}
& \makecell{\textit{full}}
& \makecell{\textit{no sign}}
& \makecell{\textit{full}}
& \makecell{\textit{no sign}} \\
\midrule

Priority (6) &
120 &
55.5 &
12.6 &
71.4 &
18.5
\\

Speed (4) &
80 &
96.3 &
17.5 &
98.8 &
18.8
\\

Obstacle (8) &
160 &
83.8 &
5.6 &
89.4 &
11.9
\\

Routing (11) &
220 &
65.6 &
23.4 &
98.2 &
50.9
\\

\midrule
\rowcolor{gray!10}
\textbf{All scenario types} &
\textbf{580} &
\textbf{72.8} &
\textbf{15.4} &
\textbf{90.3} &
\textbf{28.9}
\\

\bottomrule
\end{tabular}
\label{tab:ablation}
\end{table}

\subsection{Failure Analysis}
\label{sec:failure}


The pooled PlanT-2-FT audit separates rule violations from navigation
failures. Sign compliance is 96.6\%, but joint SCD is 71.9\%:
1,430 of 5,800 episodes (24.7\%) obey the sign but do not reach the
destination, compared with only 198 episodes (3.4\%) that violate a sign.
The remaining error is therefore dominated by route execution rather than
rule recognition.
For example, sign 4.1.4 has 100\% compliance but only 36.0\% SCD,
with 37.5\% collisions and 28.5\% off-road episodes on difficult dual-path
maps. Genuine semantic failures are concentrated in interactive rules:
crosswalk and roundabout compliance is approximately 81\%, and these two
classes account for 36.4\% of all PlanT-2-FT violation events. Privileged
experts show the same separation---94.3--98.4\% compliance but only
64.3--80.0\% SCD---confirming that their non-perfect joint scores
primarily reflect closed-loop completion failures.

\subsection{Correlation with Conventional Metrics}

\label{sec:correlation}

\begin{figure}[ht]
    \centering
    \includegraphics[width=\linewidth]{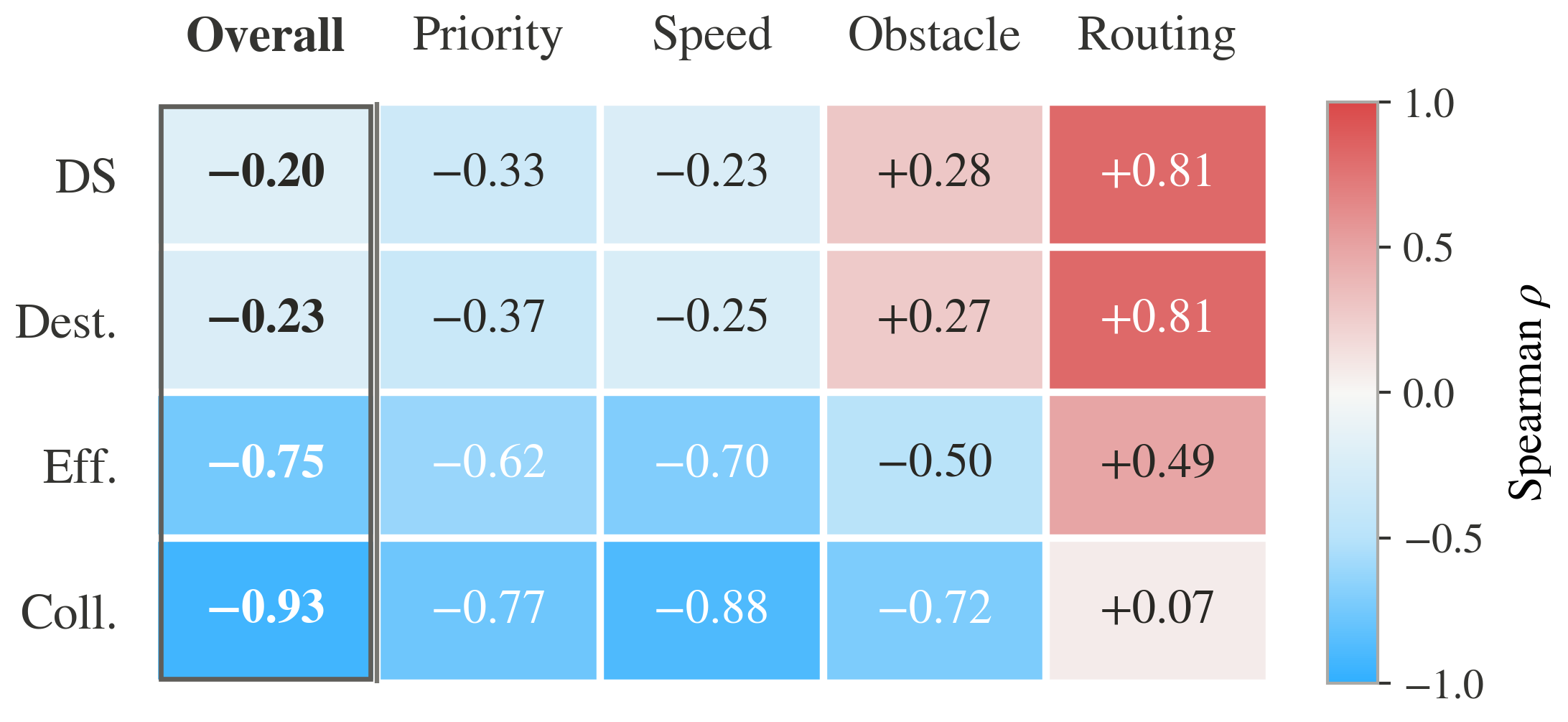}
    \caption{
    \textbf{Alignment between conventional metrics and rule-conditioned
    success.} Spearman rank correlations are computed over the 8 baseline planners and PlanT-2-FT; privileged experts are excluded.
    }
    \label{fig:correlation}
\end{figure}

Figure~\ref{fig:correlation} asks whether conventional metrics recover the
same planner ranking as SCD. We use Spearman rank correlation over the 8 non-privileged baselines and PlanT-2-FT. Since Collision Rate is lower-is-better, a negative $\rho$ indicates agreement with SCD.
No conventional metric provides a category-independent proxy for
rule-conditioned task completion. Driving Score and Destination Rate are weakly negatively associated with
overall SCD ($\rho=-0.20$ and $-0.23$, respectively), but both correlate positively with Routing SCD ($+0.81$). Six of the nine planners obtain zero
Routing SCD, so the Routing correlations are largely determined by the few
planners with non-zero rule-conditioned success in this category. Efficiency is strongly negatively associated with overall SCD ($\rho=-0.75$) and remains negative for Priority ($-0.62$), Speed ($-0.70$), and Obstacle ($-0.50$), but reverses to a positive association for Routing ($+0.49$). Collision Rate is the closest overall proxy ($\rho=-0.93$), with strong negative correlations for Priority ($-0.77$), Speed ($-0.88$), and Obstacle ($-0.72$), but almost no association with Routing ($+0.07$). These results show that the relationship between conventional metrics and
SCD varies substantially across rule categories. A planner can reach its destination and avoid collisions while still violating the target rule.
Consequently, aggregate driving metrics cannot replace explicit
rule-conditioned evaluation.

\subsection{Sign-Channel Ablation}
\label{sec:ablations}

To test whether the improvement is actually sign-conditioned,
Table~\ref{tab:ablation} replays the same PlanT-2-FT checkpoint on the same
scenes after removing sign identity
from both object tokens and the persistent sign-state token. For this ablation, we deterministically select one of the 10 variants for each test map, with different variants selected across maps.
The physical
plates remain in the scene as anonymous static objects. 
Across 580 paired
episodes, SCD falls from 72.8\% to 15.4\% ($-57.4$ points), while
compliance falls from 90.3\% to 28.9\% ($-61.4$ points). 
Every functional
group is affected, with the largest SCD drop on Obstacle scenarios
(83.8\% to 5.6\%). 
Crosswalk sign 5.19 is the sole compliance exception, remaining at 100\% in both conditions because road markings and visible
pedestrians provide a direct geometric and interactive cue. The ablation
shows that the gains cannot be attributed to generic fine-tuning alone:
explicit sign identity is necessary for the learned behavior.

\section{Conclusion}
\label{sec:conclusion}

We introduced TrafficSignBench, a rule-centric closed-loop benchmark with
automatic checkers for 34 traffic signs, 29 diverse testing scenarios, and
29,000 scenarios derived from real-map geometry and rule-targeted generation.

Standard planners achieve only 2.9--9.0\% overall SCD. Expert-supervised
fine-tuning raises PlanT-2 from 5.9\% to 72.3\% and yields 96.6\% sign
compliance, approaching the strongest privileged expert at 80.0\%. Removing
sign identity reduces SCD from 72.8\% to 15.4\%, confirming that the
improvement is explicitly rule-conditioned.

Remaining failures are primarily navigational: 24.7\% of PlanT-2-FT episodes
obey the sign but miss the destination, whereas only 3.4\% violate a sign.
The next challenge is therefore clear: preserve this measured
compliance while completing interactive right-of-way and longer-horizon routing maneuvers.

\section{Limitations}


TrafficSignBench has several limitations that define its current scope. First, we focus on traffic rules explicitly associated with traffic signs. Other regulations, such as context-dependent right-of-way rules or rules not triggered by explicit signage, are outside the current scope. 
Second, planners receive structured traffic-sign information, abstracting away sign detection and recognition errors. The benchmark therefore evaluates planning-level rule compliance rather than the complete perception-to-control pipeline.
Third, our scenarios are primarily designed around a single target rule. Although other agents and road constraints create additional interactions, the benchmark does not systematically evaluate combinations of multiple simultaneously active traffic rules. 

However, TrafficSignBench's modular design allows future extensions to be built on top, covering additional regulations or multi-rule interactions without changing the evaluation protocol.




\bibliographystyle{IEEEtran}
\bibliography{bibliography/sample-base}

\end{document}